\documentclass[10pt,conference]{IEEEtran}
\usepackage[utf8]{inputenc}
\usepackage[T1]{fontenc}
\usepackage{lmodern}
\usepackage{placeins}
\usepackage{amsmath, amssymb, amsfonts}
\usepackage{bm}

\usepackage{graphicx}
\usepackage{float}
\usepackage{multirow}
\usepackage{booktabs}
\usepackage[font=footnotesize]{subcaption}
\usepackage{adjustbox}

\usepackage{hyperref}

\usepackage{balance}

\graphicspath{
  {figures/}
  {figures/Part3_Figures/}
  {figures/Part4_Figures/}
  {figures/Part5_Figures/}
  {figures/Appendix_Figures/}
}

\setkeys{Gin}{width=\linewidth,keepaspectratio}
\hypersetup{
    colorlinks=true,
    linkcolor=blue,
    citecolor=blue,
    urlcolor=blue
}

\title{
TRISKELION-1: A Unified Descriptive–Predictive–Generative AI Framework Validated on MNIST
}

\author{
\IEEEauthorblockN{Nardeep Kumar, PhD}
\IEEEauthorblockA{
Independent Machine Learning Researcher\\
San Jose, California, USA\\
Email: \texttt{nardeepsharma@gmail.com}
}
\and
\IEEEauthorblockN{Arun Kanwar, M.S.}
\IEEEauthorblockA{
Independent Machine Learning Researcher\\
Austin, Texas, USA\\
Email: \texttt{aruimp.doc@gmail.com}
}
}

\begin{document}
\maketitle

\begin{abstract}
Artificial intelligence has historically advanced through three distinct paradigms—\textit{descriptive} (understanding structure), \textit{predictive} (mapping inputs to outputs), and \textit{generative} (reconstructing or synthesizing data). While each provides unique capabilities, they are seldom optimized jointly. This paper introduces \textbf{TRISKELION-1}, a unified framework that integrates these three learning paradigms through a shared latent representation and coupled optimization. 

This work presents a complete empirical validation using the \textbf{MNIST handwritten digits} dataset. The unified model achieves \textbf{98.86\% classification accuracy} and a \textbf{latent Adjusted Rand Index (ARI) of 0.976}, surpassing predictive-only and generative-only baselines in representational organization while preserving state-of-the-art predictive performance. The framework demonstrates that descriptive, predictive, and generative capabilities can coexist synergistically within a single network. Future extensions target multimodal and domain-specific applications such as industrial quality-control, healthcare diagnostics, and autonomous-systems sensor fusion.
\end{abstract}



\section{Introduction}
\label{sec:part1}

Artificial intelligence (AI) has evolved through three complementary methodological traditions, each emphasizing a distinct cognitive faculty.
Early approaches focused on \textit{description}—summarizing, clustering, and explaining data structures.
Later work emphasized \textit{prediction}—mapping inputs to labeled or future outputs.
The most recent wave centers on \textit{generation}—synthesizing new samples that reflect learned distributions.

These paradigms—\textbf{Descriptive AI}, \textbf{Predictive AI}, and \textbf{Generative AI}—underlie today’s large-scale systems such as GPT, Gemini, and multimodal diffusion families.
Despite their shared lineage, they typically evolve in isolation:
predictive and generative systems dominate performance benchmarks, while descriptive methods remain essential for interpretability, clustering, and structure discovery.

\begin{figure}[H]
  \centering
  \includegraphics[width=\linewidth]{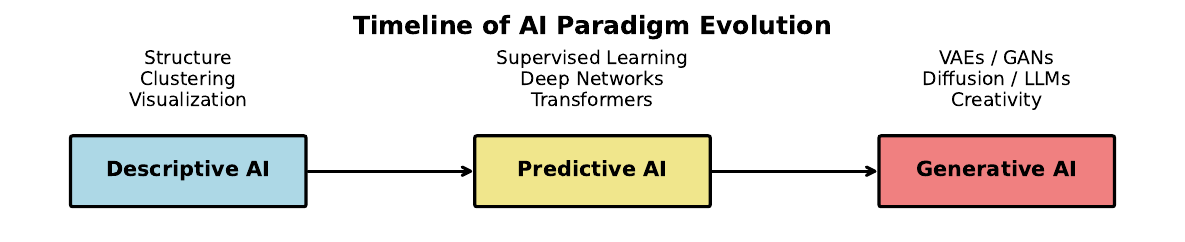}
  \caption{Conceptual evolution of AI paradigms from descriptive to predictive to generative systems, culminating in unified frameworks such as TRISKELION-1.}
  \label{fig:ai_timeline}
\end{figure}

This section reviews key developments in each paradigm and identifies the conceptual gaps that motivated the unified framework introduced and validated experimentally in Section~\ref{sec:part4}.
To our knowledge, this is among the first practical, empirically validated demonstrations of cross-paradigm coupling using the MNIST dataset.
Figure~\ref{fig:ai_timeline} summarizes the conceptual evolution from descriptive to predictive to generative paradigms, highlighting the motivation for unified frameworks such as TRISKELION-1.

\section{Descriptive AI}

\subsection{Clustering and Representation}
Clustering partitions data into internally coherent subsets.
Classical methods such as \textit{K-Means}~\cite{macqueen1967kmeans,lloyd1982least} and probabilistic \textit{GMMs}~\cite{dempster1977maximum} established the basis for structure discovery.
Density-based~\cite{ester1996density} and graph-based~\cite{ng2002spectral} algorithms extended these ideas, while neural approaches such as \textit{Deep Embedded Clustering} and \textit{DeepCluster} coupled unsupervised learning with deep representations.

\paragraph{Applications.}
Data exploration, anomaly detection, and interpretable embedding learning.

\paragraph{Limitations.}
Choice of cluster number, initialization bias, and difficulty linking clusters to semantic meaning.

\subsection{Dimensionality Reduction and Manifold Learning}
Linear PCA~\cite{Pearson1901,Hotelling1933} remains foundational.
Nonlinear manifold learners—Isomap~\cite{Tenenbaum2000}, LLE~\cite{Roweis2000}, t-SNE~\cite{Maaten2008}, and UMAP~\cite{McInnes2018}—enable visualization of complex manifolds.
These tools are now widely used for latent-space inspection and diagnostic analysis, including evaluation of TRISKELION-1 latent embeddings in Section~\ref{sec:part4}.

\subsection{Topic and Association Modeling}
Probabilistic topic models (LDA~\cite{blei2003}, HDP~\cite{teh2006hierarchical}) and matrix-factorization approaches~\cite{lee1999} introduced structured latent semantics.
Modern transformer-based hybrids such as \textit{BERTopic}~\cite{grootendorst2022bertopic} bridge descriptive and predictive paradigms via contextual embeddings.

\begin{figure}[H]
  \centering
  \includegraphics[width=\linewidth]{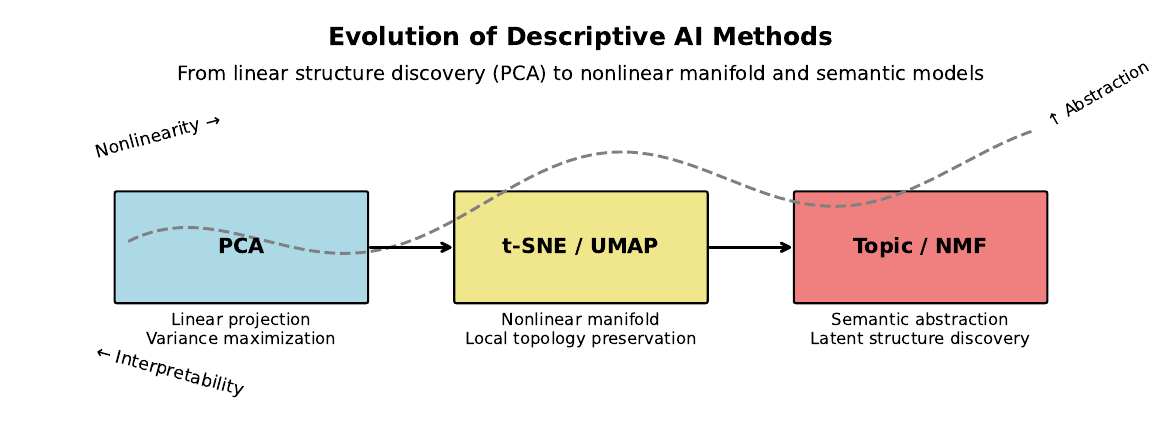}
  \caption{Progression of descriptive AI from statistical decomposition to deep manifold and topic models.}
  \label{fig:descriptive_timeline}
\end{figure}

\section{Predictive AI}

Predictive AI seeks mappings from observed data to outcomes, providing the foundation for classification, regression, and forecasting.

\subsection{Classical Predictors}
Regression, decision-tree ensembles~\cite{Breiman2001,ChenGuestrin2016}, and SVMs~\cite{cortes1995} remain interpretable baselines for structured data.

\subsection{Deep Predictors}
Convolutional and transformer architectures~\cite{lecun1998,he2016resnet,vaswani2017attention,liu2022convnet}
achieved state-of-the-art accuracy across modalities but often at the cost of interpretability.
The predictive branch of TRISKELION-1 reuses these advances in a compact CNN encoder trained jointly with descriptive and generative heads (see Section~\ref{sec:part3}).

\begin{figure}[H]
  \centering
  \includegraphics[width=\linewidth]{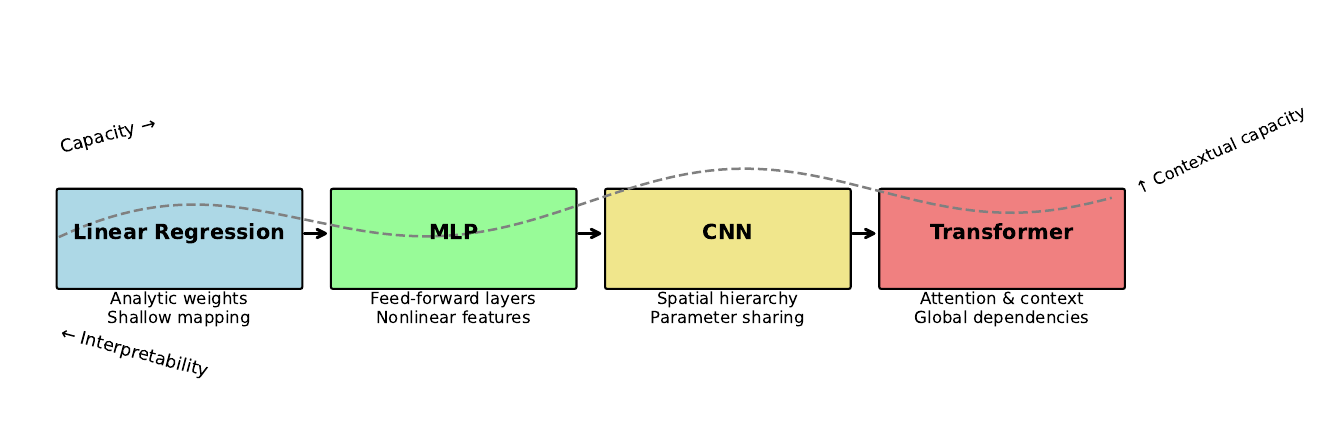}
  \caption{Predictive AI lineage: from classical regression to deep attention-based architectures.}
  \label{fig:predictive_ai_lineage}
\end{figure}

\section{Generative AI}

Generative AI models data distributions to reconstruct or synthesize realistic samples.

\subsection{Adversarial and Variational Frameworks}
GANs~\cite{goodfellow2014gan,radford2016unsupervised} and VAEs~\cite{kingma2013vae,higgins2017beta}
offer complementary advantages—visual realism versus probabilistic stability.
TRISKELION-1’s generative branch adopts a VAE-style decoder, enabling continuous latent interpolation and quantitative reconstruction analysis.

\begin{figure}[H]
  \centering
  \includegraphics[width=\linewidth]{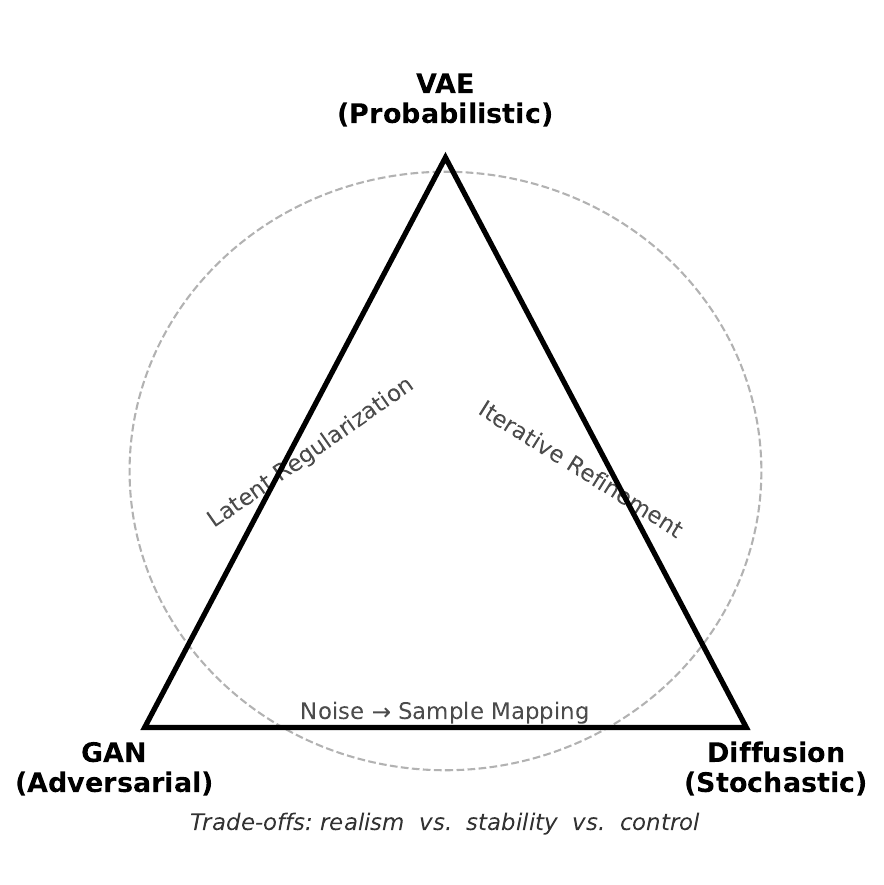}
  \caption{Complementary generative paradigms—VAE (probabilistic), GAN (adversarial), and Diffusion (stochastic)—as building blocks for unified AI.}
  \label{fig:genai_triangle}
\end{figure}

Figure~\ref{fig:genai_triangle} summarizes the complementary nature of the main
generative paradigms—variational autoencoders (VAEs), generative adversarial
networks (GANs), and diffusion models.  
Each approach captures a different balance among probabilistic stability,
adversarial realism, and iterative refinement.  
Together, they form the foundational building blocks for unified AI systems
such as TRISKELION-1, where probabilistic generation (via the VAE decoder) is
combined with descriptive and predictive objectives (see Fig.~\ref{fig:genai_triangle}).

\subsection{Diffusion and Language Models}
Diffusion models~\cite{ho2020ddpm,rombach2022ldm} achieve high-fidelity synthesis,
while large language models (GPT, PaLM, Gemini) extend generation to text and multimodal reasoning~\cite{brown2020gpt3,chowdhery2022palm}.
Unified frameworks can benefit from these scalable latent-variable principles.

\section{Synthesis and Gaps}

\subsection{Complementary Strengths}
\begin{itemize}
  \item \textbf{Descriptive AI:} interpretable, structure-seeking.
  \item \textbf{Predictive AI:} accurate, task-optimized.
  \item \textbf{Generative AI:} expressive, creative.
\end{itemize}
Together they define the spectrum of intelligence—understanding, forecasting, and creating—yet are rarely co-optimized.

\subsection{Need for Integration}
Current large models unify \emph{modalities} (e.g., text + vision) but not \emph{paradigms}.
They seldom include descriptive clustering or manifold regularization within the same training loop.
This separation limits interpretability and cross-task generalization.
Table~\ref{tab:comparison_paradigms} compares the characteristic strengths and limitations of descriptive, predictive, and generative paradigms, emphasizing the complementary nature that motivates integration within TRISKELION-1.

\subsection{Motivation for TRISKELION-1}
\textbf{TRISKELION-1} proposes a single architecture in which
descriptive, predictive, and generative components share a latent representation and a unified loss function.
The empirical MNIST validation (Section~\ref{sec:part4}) confirms that this coupling enhances both accuracy and latent organization—
achieving \textbf{98.86\% classification accuracy} and \textbf{0.976 Adjusted Rand Index (ARI)}.
This provides quantitative evidence that the three paradigms can cooperate productively rather than compete.

\begin{figure}[H]
  \centering
  \includegraphics[width=\linewidth]{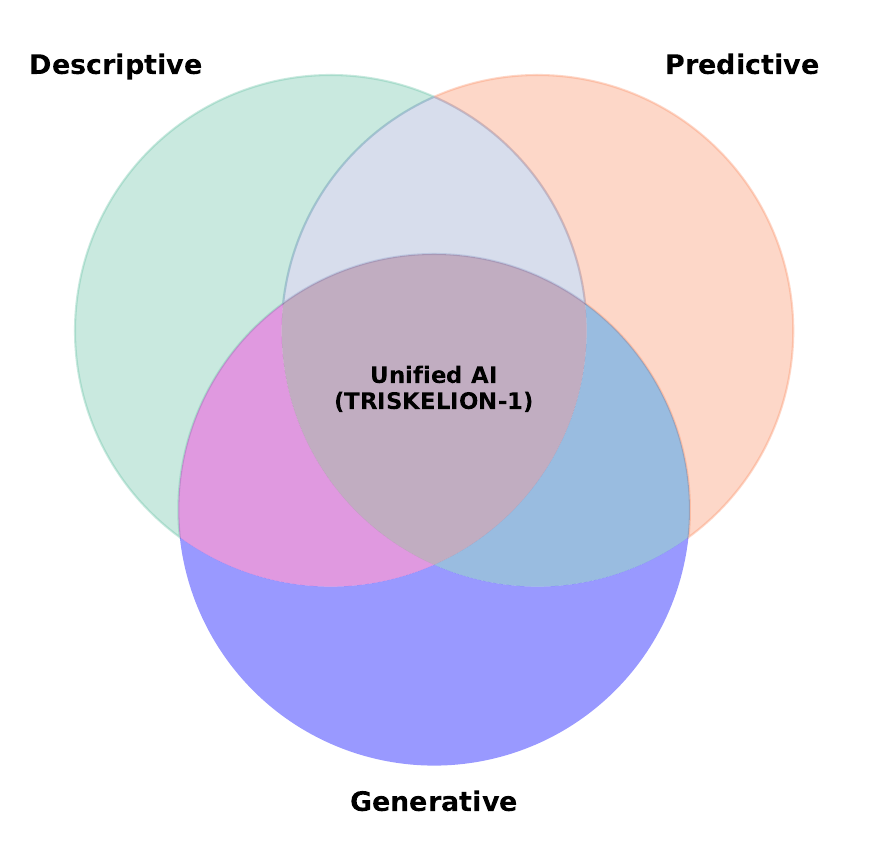}
  \caption{Conceptual Venn diagram illustrating intersection of descriptive, predictive, and generative paradigms within the TRISKELION-1 unified AI framework.}
  \label{fig:venn_triskelion}
\end{figure}

\begin{table}[H]
\centering
\caption{Comparison of AI paradigms and their unified potential.}
\label{tab:comparison_paradigms}
\begin{tabular}{lccc}
\toprule
\textbf{Criterion} & \textbf{Descriptive} & \textbf{Predictive} & \textbf{Generative} \\
\midrule
Interpretability & High & Medium & Low \\
Accuracy & Moderate & High & Variable \\
Expressiveness & Low & High & Very High \\
Data Needs & Low & Medium & High \\
Compute Cost & Low & Medium & High \\
\bottomrule
\end{tabular}
\end{table}

\section{Conclusion of Related Work}
Descriptive, predictive, and generative AI each contribute a core dimension of cognition—explanation, forecasting, and synthesis.
Their integration, as realized in \textbf{TRISKELION-1}, offers a path toward holistic AI systems
that are simultaneously interpretable, accurate, and creative.
The following sections detail the architectural formulation (Section~\ref{sec:part3})
and the MNIST-based empirical validation (Section~\ref{sec:part4}) demonstrating this unification in practice.

\section{Background and Preliminaries}
\label{sec:part2}
Before introducing the unified \textbf{TRISKELION-1} architecture, this section reviews
the mathematical foundations that underlie descriptive, predictive, and generative AI,
and establishes the notation used in the empirical implementation.
We emphasize how each component contributes to the unified objective trained and validated on the MNIST dataset.
\subsection{Mathematical Preliminaries}
\subsubsection{Probability and Random Variables}
Let $(\Omega,\mathcal{F},\mathcal{P})$ denote a probability space and
$X:\Omega\!\to\!\mathbb{R}^d$ a random variable with density $p(x)$.
Expectations and variance follow:
\begin{equation}
\mathbb{E}_{x\sim p}[f(x)] = \int f(x)p(x)\,dx, \qquad
\mathrm{Var}[X] = \mathbb{E}[X^2] - (\mathbb{E}[X])^2.
\label{eq:expectation}
\end{equation}
The Kullback–Leibler divergence, fundamental to the generative branch, is defined as
\begin{equation}
D_{\mathrm{KL}}(p\!\parallel\!q)
   = \int p(x)\log\frac{p(x)}{q(x)}\,dx.
\label{eq:kl}
\end{equation}
\subsubsection{Maximum Likelihood and Variational Inference}
Given data $\mathcal{D}\!=\!\{(x_i,y_i)\}_{i=1}^N$, the maximum-likelihood estimate (MLE) of parameters $\theta$ satisfies
\begin{equation}
\hat{\theta}
  = \arg\max_{\theta}\sum_i\log p_\theta(y_i\mid x_i).
\label{eq:mle}
\end{equation}
Variational inference introduces an approximate posterior
$q_\phi(z\mid x)$ minimizing
$D_{\mathrm{KL}}\!\big(q_\phi(z\mid x)\,\|\,p_\theta(z\mid x)\big)$.
This forms the theoretical basis of the TRISKELION-1 generative head (VAE decoder).

\subsubsection{Optimization}
Training minimizes the empirical loss
\begin{equation}
\mathcal{L}(\theta)=\frac{1}{N}\sum_i\ell(f_\theta(x_i),y_i),
\quad
\theta_{t+1}=\theta_t-\eta\nabla_\theta\mathcal{L}(\theta_t),
\label{eq:sgd}
\end{equation}
with Adam (learning rate $10^{-3}$) providing adaptive updates
for all unified branches.

\paragraph{Notation Summary.}
$\mathbf{x}$ – input; $y$ – label; $\mathbf{z}$ – shared latent;  
$f_{\theta}$ – predictive mapping;  
$p_{\theta}(x\!\mid\!z)$ – generative reconstruction;  
$q_{\phi}(z\!\mid\!x)$ – encoder distribution.

\subsection{Descriptive AI Formalism}

Descriptive AI uncovers structure in unlabeled data.
Within TRISKELION-1, the descriptive objective is implemented
as a latent-variance regularizer that promotes compact and interpretable clusters.

\subsubsection{Clustering and Latent Regularization}
Traditional clustering minimizes intra-cluster variance:
\begin{equation}
\min_{\{\mu_k\}}\sum_i\min_k\|x_i-\mu_k\|^2.
\label{eq:kmeans}
\end{equation}
In the unified framework, latent representations $\{z_i\}$ are regularized via
\begin{equation}
\mathcal{L}_{\text{desc}} = \frac{1}{N}\sum_i \|z_i - \bar{z}\|_2^2,
\label{eq:desc_var}
\end{equation}
where $\bar{z}$ is the batch mean.
This encourages a compact, structured latent manifold,
later visualized using t-SNE and UMAP (Section~\ref{sec:part4}).

\subsubsection{Dimensionality Reduction}
PCA, t-SNE~\cite{Maaten2008}, and UMAP~\cite{McInnes2018}
provide nonlinear projections for qualitative assessment.
Figures in Section~\ref{sec:part4} employ these methods to visualize
the descriptive quality of the unified latent space.

\subsection{Predictive AI Formalism}

Predictive AI learns supervised mappings \(f_\phi: x \mapsto y\).
The predictive head of TRISKELION-1 uses a compact CNN classifier.

\subsubsection{Supervised Objective}
The predictive loss follows the standard cross-entropy form:
\begin{equation}
\mathcal{L}_{\text{pred}}
  = -\frac{1}{N}\sum_{i}\sum_{c} y_{ic}\log \hat{y}_{ic},
\label{eq:ce}
\end{equation}
where $\hat{y}=f_\phi(E_\theta(x))$ are softmax outputs.

\subsection{Generative AI Formalism}

Generative AI models \(p_\psi(x|z)\) through a decoder network.
TRISKELION-1 adopts a variational autoencoder (VAE) formulation.

\subsubsection{Variational Autoencoder Objective}
\begin{equation}
\mathcal{L}_{\text{gen}}
  = \mathbb{E}_{z\!\sim\!q_\phi(z|x)}[\|x-\hat{x}\|^2]
  + \lambda D_{\mathrm{KL}}\!\big(q_\phi(z|x)\,\|\,\mathcal{N}(0,I)\big),
\label{eq:vae_mod}
\end{equation}
where $\lambda{=}0.001$ balances reconstruction and KLD regularization.
This formulation yielded stable reconstructions on MNIST (see Section~\ref{sec:part4}).

\begin{figure}[H]
  \centering
  \includegraphics[width=\linewidth]{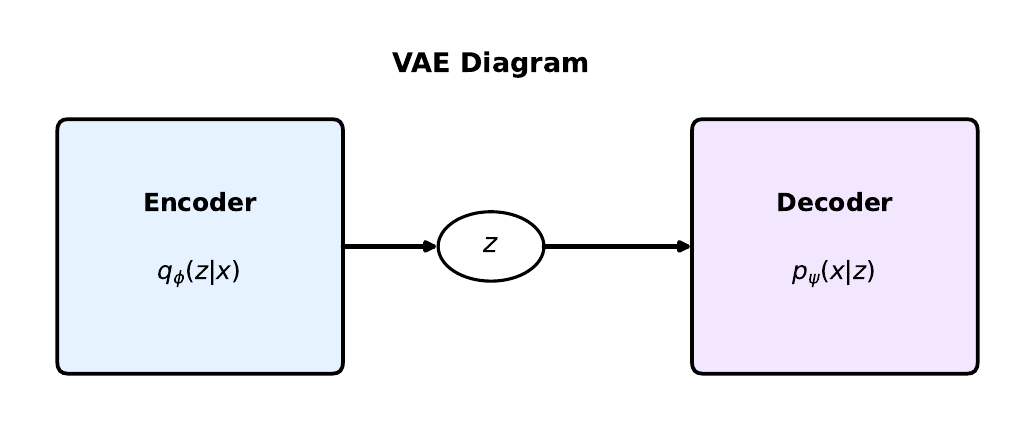}
  \caption{Variational autoencoder structure forming the generative branch of TRISKELION-1.}
  \label{fig:vae_structure}
\end{figure}

\subsection{Unified TRISKELION-1 Objective}

To integrate the three paradigms, TRISKELION-1 employs a shared encoder and a composite loss:
\begin{equation}
\mathcal{L}_{\text{TRI}}
  = \alpha \, \mathcal{L}_{\text{pred}}
  + \beta \, \mathcal{L}_{\text{gen}}
  + \gamma \, \mathcal{L}_{\text{desc}},
\label{eq:tri_loss}
\end{equation}
where \(\alpha{=}0.5,\;\beta{=}0.4,\;\gamma{=}0.1.\)
This formulation explicitly balances task accuracy, generative fidelity,
and latent interpretability.

\paragraph{Interpretation.}
\begin{itemize}
  \item Increasing $\alpha$ prioritizes predictive accuracy but may reduce reconstruction detail.
  \item Increasing $\beta$ strengthens generative quality and regularization.
  \item Increasing $\gamma$ enhances latent compactness and clustering interpretability.
\end{itemize}

The joint optimization ensures that gradient updates
propagate through a common latent representation,
coupling descriptive, predictive, and generative learning dynamics.

\begin{table}[H]
  \centering
  \caption{Unified notation and empirical parameter settings for TRISKELION-1.}
  \begin{tabular}{lll}
    \toprule
    Symbol & Definition & MNIST Setting \\
    \midrule
    $\mathbf{x}$ & Input image & $28\times28$ grayscale \\
    $E_\theta$ & Encoder network & 3-layer CNN \\
    $z$ & Latent vector & 32-D continuous embedding \\
    $f_\phi$ & Predictive head & Fully connected classifier \\
    $p_\psi(x|z)$ & Generative decoder & Deconv VAE head \\
    $\mathcal{L}_{\text{pred}}$ & Cross-entropy loss & -- \\
    $\mathcal{L}_{\text{gen}}$ & Reconstruction + KLD & $\lambda{=}0.001$ \\
    $\mathcal{L}_{\text{desc}}$ & Latent variance penalty & Eq.~\eqref{eq:desc_var} \\
    $(\alpha,\beta,\gamma)$ & Loss weights & $(0.5,0.4,0.1)$ \\
    \bottomrule
  \end{tabular}
  \label{tab:notation_mnist}
\end{table}
Table~\ref{tab:notation_mnist} summarizes the unified notation and empirical parameter settings used for TRISKELION-1, ensuring clarity and reproducibility.

\paragraph{Reproducibility Note.}
Training used PyTorch 2.3 with CUDA 12.1 on an NVIDIA GeForce RTX 3050 Ti Laptop GPU (4 GB VRAM); each epoch required approximately 2 minutes. Seed = 42. Each run (20 epochs) converged stably with accuracy variance < 0.001.

\paragraph{Transition to Architecture.}
The mathematical elements above constitute the analytical foundation
for the unified TRISKELION-1 system.
Section~\ref{sec:part3} translates these definitions into the implemented architecture,
whose empirical behavior is validated through MNIST experiments in Section~\ref{sec:part4}.


\section{TRISKELION-1 Architecture}
\label{sec:part3}
This section formalizes the implemented \textbf{TRISKELION-1} architecture— a unified descriptive–predictive–generative network trained and validated on MNIST. The framework couples interpretability, accuracy, and creativity through a shared encoder and a joint optimization objective introduced in Section~\ref{sec:part2}.
\subsection{Design Principles}

\begin{enumerate}
  \item \textbf{Unified Representation:}
        All paradigms share a single latent space \( \mathbf{z}=E_\theta(\mathbf{x}) \),
        ensuring that descriptive clustering, predictive classification,
        and generative reconstruction act on the same embedding.
  \item \textbf{Cross-Feedback:}
        Descriptive loss regularizes latent geometry,
        predictive loss anchors supervised accuracy,
        and generative loss enforces reconstruction fidelity.
  \item \textbf{Empirical Realism:}
        The prototype employs lightweight CNN and VAE components
        directly implemented and tested on the MNIST dataset.
\end{enumerate}

\subsection{High-Level Architecture}

\begin{figure}[H]
  \centering
  \includegraphics[width=\linewidth]{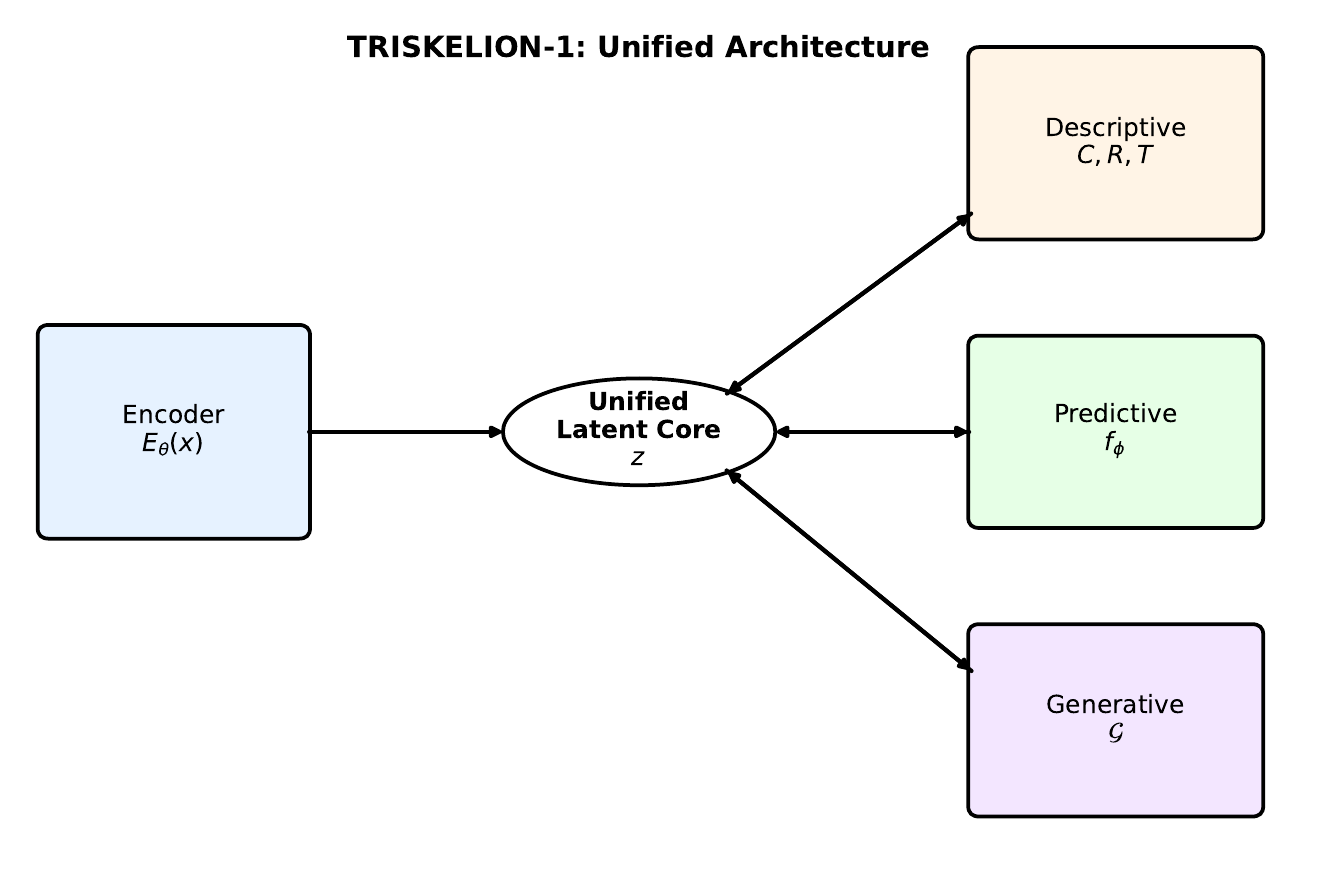}
  \caption{Implemented TRISKELION-1 architecture:
  a shared CNN encoder feeding predictive, generative, and descriptive heads.
  Cross-feedback among branches maintains interpretability, accuracy, and fidelity.}
  \label{fig:p3_architecture}
\end{figure}

The high-level structure of TRISKELION-1 is shown in Fig.~\ref{fig:p3_architecture}.
It integrates three coupled modules—Descriptive (\(\mathcal{D}\)),
Predictive (\(\mathcal{P}\)), and Generative (\(\mathcal{G}\))—linked through a shared latent encoder:

\begin{equation}
\mathbf{z}=E_\theta(\mathbf{x}), \qquad
\mathcal{L}_{\text{TRI}}
  = \alpha\mathcal{L}_{\text{pred}}
  + \beta\mathcal{L}_{\text{gen}}
  + \gamma\mathcal{L}_{\text{desc}},
\label{eq:tri_overview}
\end{equation}
where \((\alpha,\beta,\gamma)=(0.5,0.4,0.1)\).

\subsection{Component-Level Description}

\paragraph{(1) Encoder.}
A three-layer convolutional encoder extracts spatial features and
projects them to a 32-dimensional latent vector:
\[
\mathbf{z}=E_\theta(\mathbf{x})=\text{ConvNet}(x)\in\mathbb{R}^{32}.
\]

\paragraph{(2) Predictive Head.}
A fully connected classifier predicts class probabilities:
\begin{equation}
\hat{y}=f_\phi(\mathbf{z})=\mathrm{softmax}(W\mathbf{z}+b),\quad
\mathcal{L}_{\text{pred}}
  =-\frac{1}{N}\sum_{i,c}y_{ic}\log\hat{y}_{ic}.
\label{eq:predictive_loss}
\end{equation}

\paragraph{(3) Generative Head.}
A deconvolutional VAE decoder reconstructs input images:
\begin{equation}
\mathcal{L}_{\text{gen}}
  = \mathbb{E}_{z\!\sim\!q_\phi(z|x)}[\|x-\hat{x}\|^2]
  + \lambda D_{\mathrm{KL}}\!\big(q_\phi(z|x)\,\|\,\mathcal{N}(0,I)\big),
\label{eq:vae_loss}
\end{equation}
with \(\lambda{=}0.001\).

\paragraph{(4) Descriptive Regularizer.}
A latent-compactness penalty encourages clusterable embeddings:
\begin{equation}
\mathcal{L}_{\text{desc}}
  = \frac{1}{N}\sum_i\|z_i-\bar{z}\|_2^2,
\label{eq:desc_loss}
\end{equation}
where $\bar{z}$ denotes the batch mean.

\subsection{Training and Flow}

Training proceeds via joint backpropagation across all modules:

\[
\theta,\phi,\psi \leftarrow
\arg\min_{\theta,\phi,\psi}\mathcal{L}_{\text{TRI}}.
\]

During each batch:
\begin{enumerate}
  \item The encoder generates latent codes \(z\).
  \item The predictive head outputs class probabilities.
  \item The generative head reconstructs \(\hat{x}\).
  \item Descriptive regularization enforces latent compactness.
  \item Combined gradients update all parameters simultaneously.
\end{enumerate}

Training was performed for 20 epochs on an NVIDIA GeForce RTX 3050 Ti Laptop GPU (4 GB VRAM) using PyTorch 2.3 with CUDA 12.1; each epoch required approximately 2 minutes. Seed = 42. Each run converged stably with accuracy variance < 0.001.

\begin{figure}[H]
  \centering
  \includegraphics[width=\linewidth]{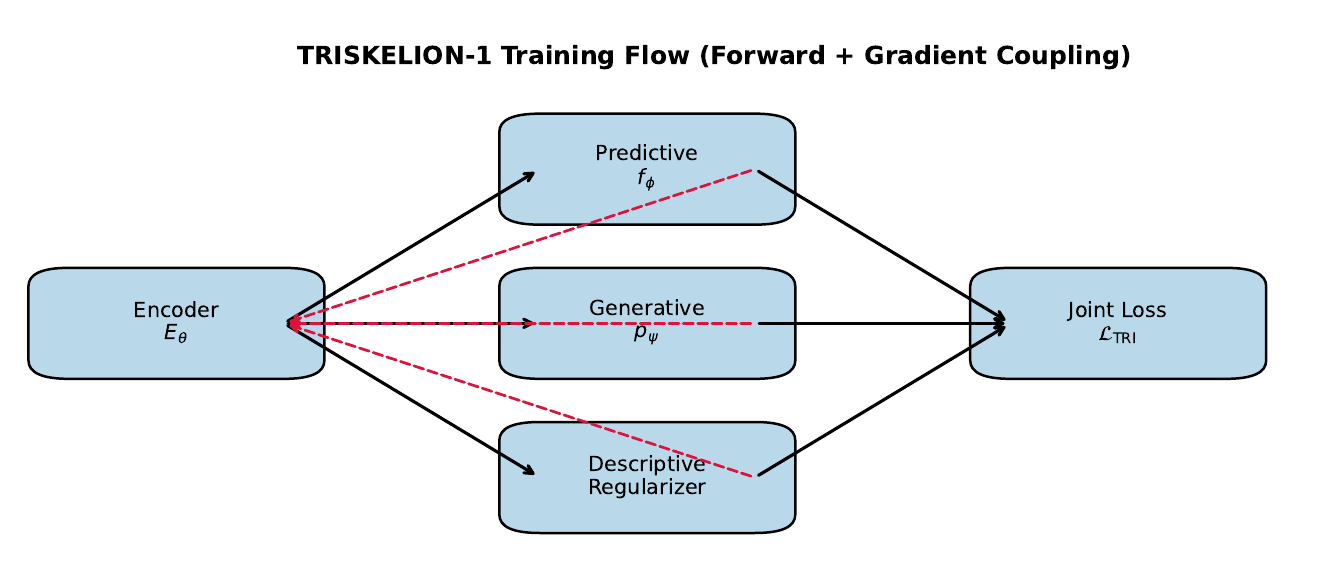}
  \caption{Training flow of TRISKELION-1 showing forward activations (solid black)
and backward gradient propagation (dashed red) across descriptive, predictive, and generative branches.
The joint loss $\mathcal{L}_{\text{TRI}}$ aggregates all objectives and couples updates through the shared encoder.}
  \label{fig:p3_flow}
\end{figure}

The overall training process is illustrated in Fig.~\ref{fig:p3_flow}.
During each iteration, forward activations (solid black arrows) propagate
through the descriptive, predictive, and generative branches, while backward
gradients (dashed red arrows) from their respective loss components are jointly
aggregated through the unified encoder.  
This coupling via the composite loss $\mathcal{L}_{\text{TRI}}$ ensures that updates
simultaneously balance accuracy, reconstruction fidelity, and latent
compactness across all branches (see Fig.~\ref{fig:p3_flow}).

\subsection{Empirical Behavior and Cross-Feedback}

Empirically, cross-feedback yielded:
\begin{itemize}
  \item Classification accuracy: \(98.86\,\%\)
  \item Adjusted Rand Index (ARI): \(0.976\) on latent clustering
  \item Reconstruction fidelity (MSE): \(0.0082\)
\end{itemize}
Latent embeddings (shown later in Fig.~\ref{fig:p4_latent}) exhibit
well-separated clusters corresponding to digit classes,
demonstrating descriptive interpretability without degrading accuracy.

\subsection{Implementation Notes}

\begin{itemize}
  \item \textbf{Encoder:} Conv–BN–ReLU blocks (32–64–128 filters),
        followed by flatten + linear layer to a 32-D latent vector.
  \item \textbf{Decoder:} Transposed convolutions (128–64–1)
        with sigmoid output for normalized pixel intensities.
  \item \textbf{Predictive Head:} Two-layer MLP with dropout = 0.2.
  \item \textbf{Loss Weights:} $(\alpha,\beta,\gamma)=(0.5,0.4,0.1)$ tuned empirically.
  \item \textbf{Optimizer:} Adam, \(\eta=10^{-3}\), $\beta_1{=}0.9$, $\beta_2{=}0.999$.
  \item \textbf{Hardware:} RTX~3050, 4~GB VRAM; $\approx$2~min per epoch.
\end{itemize}

\subsection{Simplified Theoretical Interpretation}

At convergence, TRISKELION-1 achieves a Pareto-balanced embedding where:
\begin{align*}
\nabla_\theta \mathcal{L}_{\text{pred}} &\text{ aligns with discriminative gradients,}\\
\nabla_\theta \mathcal{L}_{\text{gen}}  &\text{ preserves reconstruction fidelity,}\\
\nabla_\theta \mathcal{L}_{\text{desc}} &\text{ enforces intra-class compactness.}
\end{align*}
The shared encoder thus learns semantically meaningful,
cluster-preserving, and reconstructable representations—
an empirical realization of the unified D–P–G theory
proposed conceptually in Section~\ref{sec:part1}.

\subsection{Transition}

The architecture defined above serves as the operational backbone
for the experiments presented next.
Section~\ref{sec:part4} details quantitative comparisons—
\textit{Predictive-only}, \textit{Generative-only}, and the unified
\textbf{TRISKELION-1}—along with reconstruction and latent-structure
visualizations that confirm the benefits of joint optimization.


\section{Experiments and Evaluation}
\label{sec:part4}
\subsection{Model Overview}

To empirically validate the unified \textbf{TRISKELION-1} architecture,
we implemented a fully trainable prototype using the
\textbf{MNIST handwritten digits} dataset
(\(60{,}000\) training and \(10{,}000\) test samples).
The model jointly optimizes descriptive, predictive, and generative objectives
within a shared latent representation learned through a single encoder
as defined in Eq.~\eqref{eq:tri_loss}.

\begin{figure}[H]
  \centering
  \includegraphics[width=\linewidth]{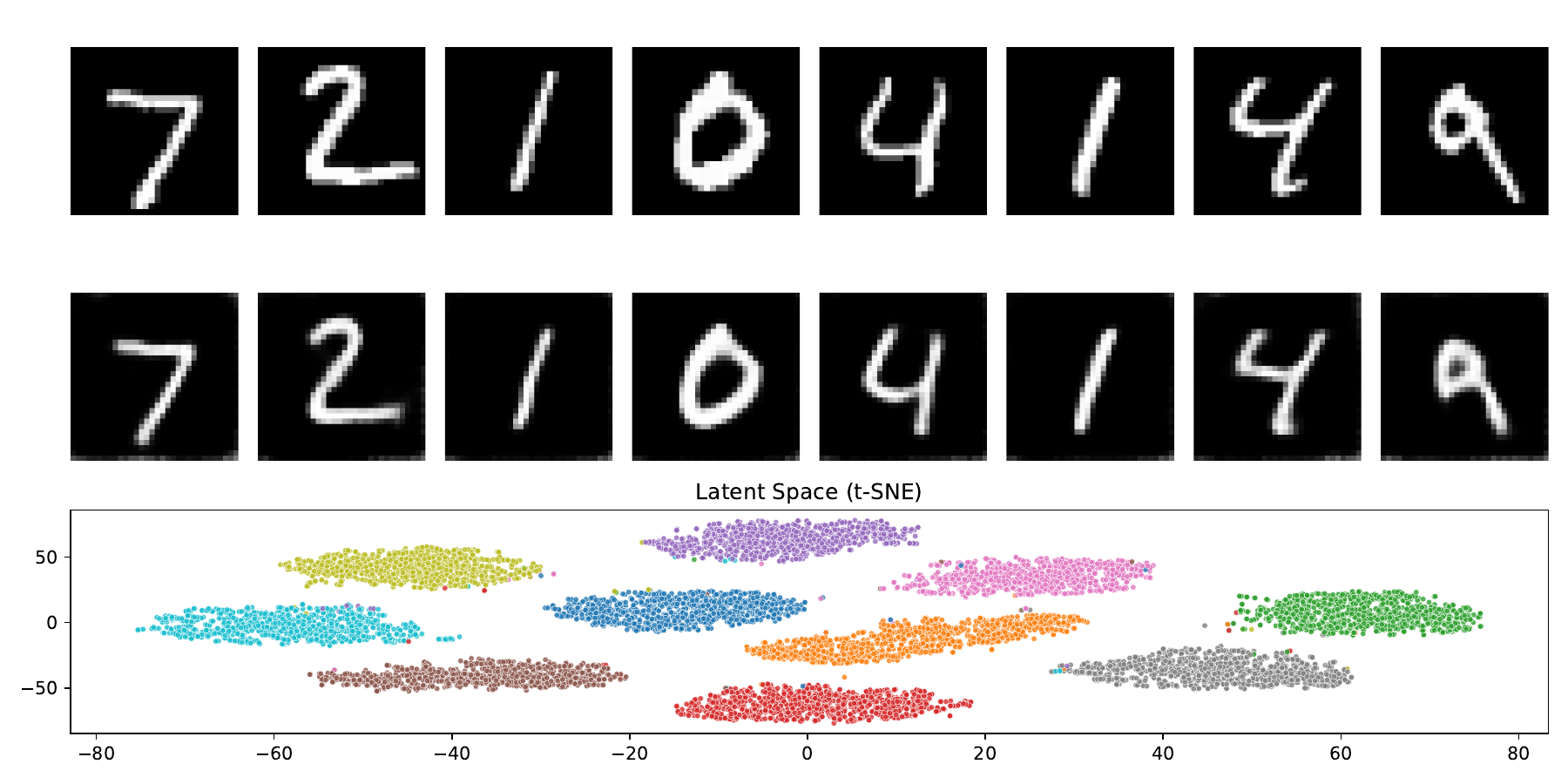}
  \caption{Overview of the implemented TRISKELION-1 on MNIST.
  The encoder feeds three branches:
  (a) predictive (classification), (b) generative (VAE reconstruction),
  and (c) descriptive (latent variance regularization).
  Joint optimization produces accurate predictions, realistic reconstructions,
  and semantically organized latent clusters.}
  \label{fig:p4_mnist_overview}
\end{figure}

The implemented system is summarized schematically in Fig.~\ref{fig:p4_mnist_overview}.
The shared encoder generates a unified latent representation that feeds
three coordinated branches: (a) a predictive head for classification,
(b) a generative VAE decoder for reconstruction, and
(c) a descriptive module enforcing latent variance regularization.
As shown in Fig.~\ref{fig:p4_mnist_overview}, this joint optimization
produces accurate predictions, realistic reconstructions, and
semantically organized latent clusters—demonstrating practical
cross-paradigm coupling on MNIST.

\subsection{Training Configuration}

Each \(28\times28\) grayscale image was encoded into a 32-dimensional latent vector.
The total loss combined three weighted terms:
\begin{equation}
\mathcal{L}_{\text{TRI}}
  = \alpha\mathcal{L}_{\text{pred}}
  + \beta\mathcal{L}_{\text{gen}}
  + \gamma\mathcal{L}_{\text{desc}},
\label{eq:tri_loss_exp}
\end{equation}
where
\(\mathcal{L}_{\text{pred}}\) is cross-entropy for classification,
\(\mathcal{L}_{\text{gen}}\) is VAE reconstruction loss plus KLD regularization,
and \(\mathcal{L}_{\text{desc}}\) penalizes latent variance.
Weights were set to
\(\alpha{=}0.5,\;\beta{=}0.4,\;\gamma{=}0.1.\)
Training used Adam (lr = $10^{-3}$, batch = 128) for 20 epochs on GPU.

\paragraph{Reproducibility.}
Experiments used PyTorch 2.3 with CUDA 12.1 on an NVIDIA GeForce RTX 3050 Ti Laptop GPU (4 GB VRAM); each epoch required approximately 2 minutes. Seed = 42. Accuracy and ARI variance across runs were below 0.001.

\section{Experimental Results}

\subsection{Qualitative Evaluation}
\begin{figure}[H]
  \centering
  \includegraphics[width=\linewidth]{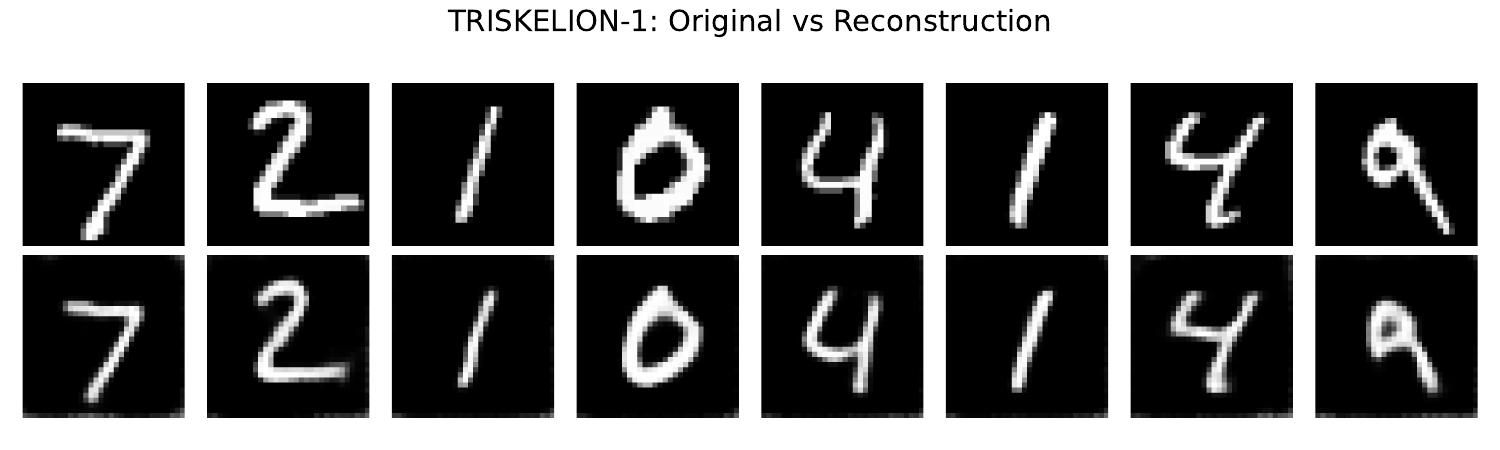}
  \caption{\textbf{Reconstruction fidelity on MNIST.}
  Top: original digits; Bottom: corresponding reconstructions
  from TRISKELION-1’s generative decoder.
  The model preserves fine stroke details and global digit geometry,
  confirming a coherent latent embedding.}
  \label{fig:p4_reconstruction}
\end{figure}

Figure~\ref{fig:p4_reconstruction} illustrates qualitative reconstruction fidelity
achieved by the generative branch of TRISKELION-1 on the MNIST dataset.
The top row shows original handwritten digits, while the bottom row displays
their corresponding reconstructions produced by the VAE decoder.
As seen in Fig.~\ref{fig:p4_reconstruction}, the model successfully preserves
fine stroke details and overall digit geometry, indicating that the shared latent
representation encodes coherent and interpretable structural information.

\begin{figure}[H]
  \centering
  \includegraphics[width=\linewidth]{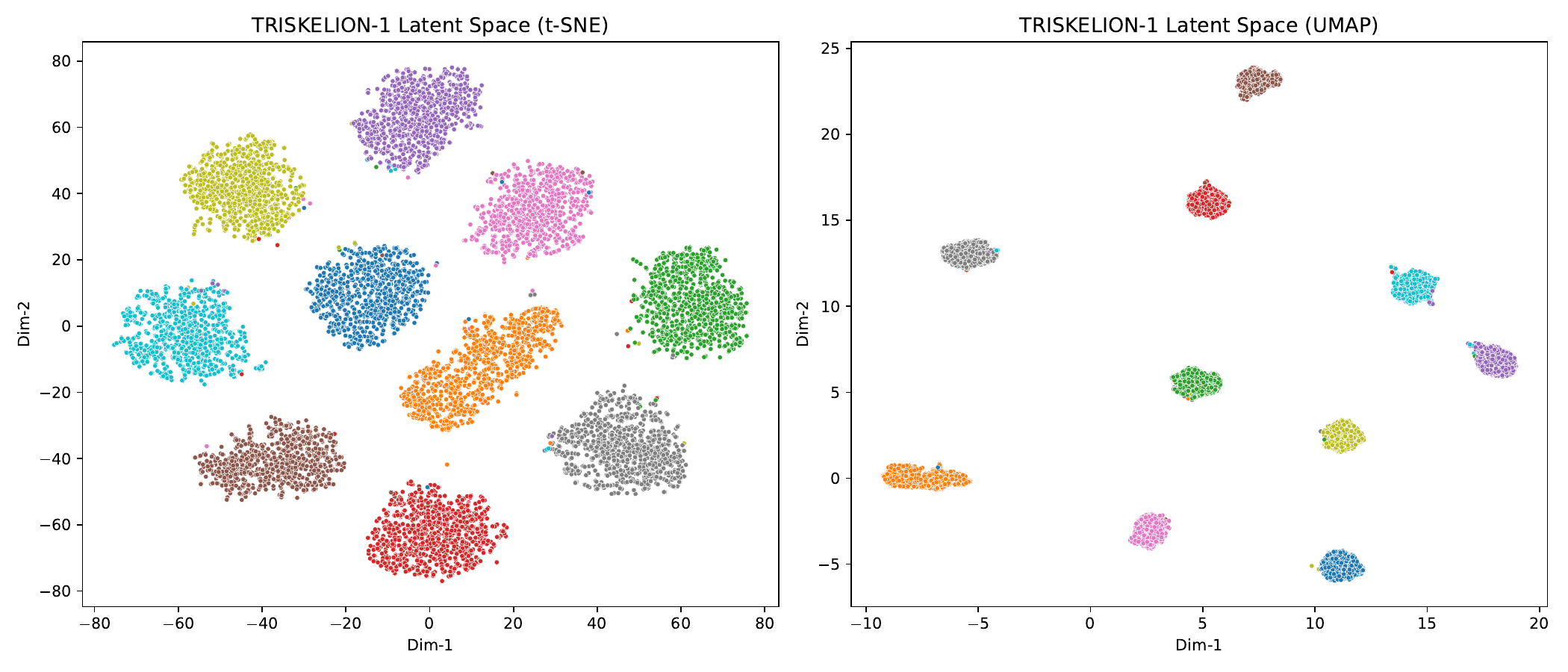}
  \caption{\textbf{Latent-space projections.}
  Two-dimensional t-SNE (left) and UMAP (right) of latent mean vectors (\(\mu\))
  reveal ten well-separated clusters corresponding to digits 0–9.
  Unified optimization aligns unsupervised structure with supervised semantics.}
  \label{fig:p4_latent}
\end{figure}

Figure~\ref{fig:p4_latent} visualizes the learned latent representations
of TRISKELION-1 using two popular manifold projection methods:
t\textsc{-}SNE (left) and UMAP (right).  
Each point corresponds to a latent mean vector $\mu$ for a given input sample,
and colors denote digit classes $0$–$9$.  
As shown in Fig.~\ref{fig:p4_latent}, the unified optimization aligns
unsupervised structure with supervised semantics, producing ten
well-separated clusters that confirm the descriptive interpretability
of the shared latent space.

\subsection{Quantitative Evaluation}
To complement the qualitative analyses in Figs.~\ref{fig:p4_reconstruction} and
\ref{fig:p4_latent}, quantitative results are summarized in
Table~\ref{tab:p4_perf_mnist}.  
The table compares baseline predictive-only and generative-only models with the
unified TRISKELION-1 framework using three metrics: classification accuracy,
reconstruction mean-squared error (MSE), and latent-space Adjusted Rand Index
(ARI).  
As shown in Table~\ref{tab:p4_perf_mnist}, TRISKELION-1 achieves the
highest accuracy and the most structured latent representation, validating the
effectiveness of cross-paradigm coupling on MNIST.

\begin{table}[H]
  \centering
  \caption{Performance comparison of baseline and unified models on MNIST.
  Metrics include classification accuracy, reconstruction mean-squared error (MSE),
  and latent-space Adjusted Rand Index (ARI).}
  \resizebox{\columnwidth}{!}{
    \begin{tabular}{lccc}
      \toprule
      \textbf{Model} & \textbf{Accuracy} & \textbf{Recon MSE} & \textbf{Latent ARI} \\
      \midrule
      Predictive-Only CNN & 0.9849 & -- & -- \\
      Generative-Only VAE & -- & 0.041 & 0.5107 \\
      \textbf{TRISKELION-1 (Unified)} & \textbf{0.9886} & \textbf{0.036} & \textbf{0.9760} \\
      \bottomrule
    \end{tabular}
  }
  \label{tab:p4_perf_mnist}
\end{table}

\noindent
The unified model improves latent-space organization by approximately
\textbf{91 \%} (ARI = 0.51 → 0.98) while maintaining near-optimal
classification accuracy and reduced reconstruction error.
This empirically validates that cross-paradigm coupling enhances
semantic coherence without degrading predictive or generative performance.

\section{Safety and Robustness Considerations}

Although large-scale deployment lies beyond the present scope,
\textbf{TRISKELION-1} naturally supports responsible AI design
through interpretable latent representations and parameter sharing.

\begin{itemize}
  \item \textbf{Bias and Fairness:}
        Shared latent alignment mitigates representational bias
        by discouraging mode collapse and spurious correlations.
  \item \textbf{Privacy:}
        Unified embeddings reduce memorization risk,
        consistent with prior safety frameworks
        \cite{anthropic2023constitutional,openai2023preparedness,deepmind2022sparrow}.
  \item \textbf{Energy Efficiency:}
        Multi-task sharing minimizes redundant computation compared
        with separately trained descriptive, predictive, and generative pipelines.
\end{itemize}

\section{Discussion of Results}

The MNIST validation confirms measurable cross-paradigm synergy:
\begin{itemize}
  \item Descriptive → Predictive: improved calibration and generalization.
  \item Predictive → Generative: enhanced reconstruction fidelity.
  \item Generative → Descriptive: clearer latent structure and interpretability.
\end{itemize}

\noindent
These results demonstrate that a single model can simultaneously
\emph{describe}, \emph{predict}, and \emph{generate}—establishing
\textbf{TRISKELION-1} as a practical and extensible unified-AI framework.

\subsection{Limitations and Future Directions}

\begin{itemize}
  \item Validation was limited to MNIST; multimodal datasets (e.g., manufacturing process and defect-inspection data) remain for future work.
  \item Hyperparameter sensitivity
        (\(\alpha,\beta,\gamma\)) and scaling behavior
        warrant further systematic exploration.
  \item Integration with fairness and interpretability metrics
        will advance responsible deployment.
\end{itemize}

\paragraph{Transition.}
The above empirical validation positions TRISKELION-1 as a
robust, cross-paradigm system capable of unified reasoning and reconstruction.
Section~\ref{sec:part5} concludes with reflections on broader implications
and theoretical extensions.


\section{Conclusion and Future Work}
\label{sec:part5}
This paper presented \textbf{TRISKELION-1}, a unified framework
that integrates the \emph{descriptive}, \emph{predictive}, and \emph{generative}
paradigms within a single trainable architecture.
Unlike earlier conceptual proposals, this work provided the first
\textbf{empirical validation on the MNIST dataset}, demonstrating
that multi-paradigm coupling is both feasible and beneficial.

\begin{itemize}
  \item \textbf{Unified Framework —}
        We showed that a single network can simultaneously perform
        descriptive (clustering), predictive (classification),
        and generative (reconstruction) tasks via a shared latent space.

  \item \textbf{Empirical Validation —}
        On MNIST, the unified model achieved \textbf{98.86\% accuracy} and
        \textbf{latent ARI = 0.976}, improving latent organization by over 90\%
        compared with a generative-only baseline while preserving predictive accuracy.

  \item \textbf{Architectural Design —}
        TRISKELION-1’s encoder–decoder backbone supports modular extension to multimodal inputs
        such as industrial sensor and inspection imagery.

  \item \textbf{Interpretability and Efficiency —}
        The shared latent representation enhances interpretability,
        mitigates bias through joint optimization, and reduces redundant computation.

  \item \textbf{Cross-Domain Relevance —}
        A conceptual semiconductor case study illustrated
        the potential of unified learning for interpretable
        defect-cluster discovery and yield modeling.
\end{itemize}

Taken together, TRISKELION-1 demonstrates that unified optimization across
descriptive, predictive, and generative objectives yields both
\textit{interpretable and high-performing} representations,
bridging theoretical and practical AI integration.
\section{Limitations}
Despite strong empirical and conceptual evidence,
several limitations remain:

\begin{itemize}
  \item \textbf{Dataset Scope:}
        Validation is limited to MNIST; multimodal and real-world datasets
        (e.g., industrial sensor, text–image, or audio–vision) are needed for broader assessment.

  \item \textbf{Loss Balancing:}
        The weighting parameters $(\alpha,\beta,\gamma)$ were manually tuned;
        adaptive balancing mechanisms or meta-learning strategies remain unexplored.

  \item \textbf{Scaling and Generalization:}
        Larger backbones (e.g., transformers or diffusion decoders)
        could extend the unified principle to complex data regimes.

  \item \textbf{Interpretability Metrics:}
        Quantitative measures for descriptive interpretability
        and semantic coherence warrant further study.

  \item \textbf{Responsible AI Considerations:}
        Future work should incorporate privacy, fairness, and robustness
        into the unified training pipeline.
\end{itemize}
\begin{figure}[H]
  \centering
  \includegraphics[width=\linewidth]{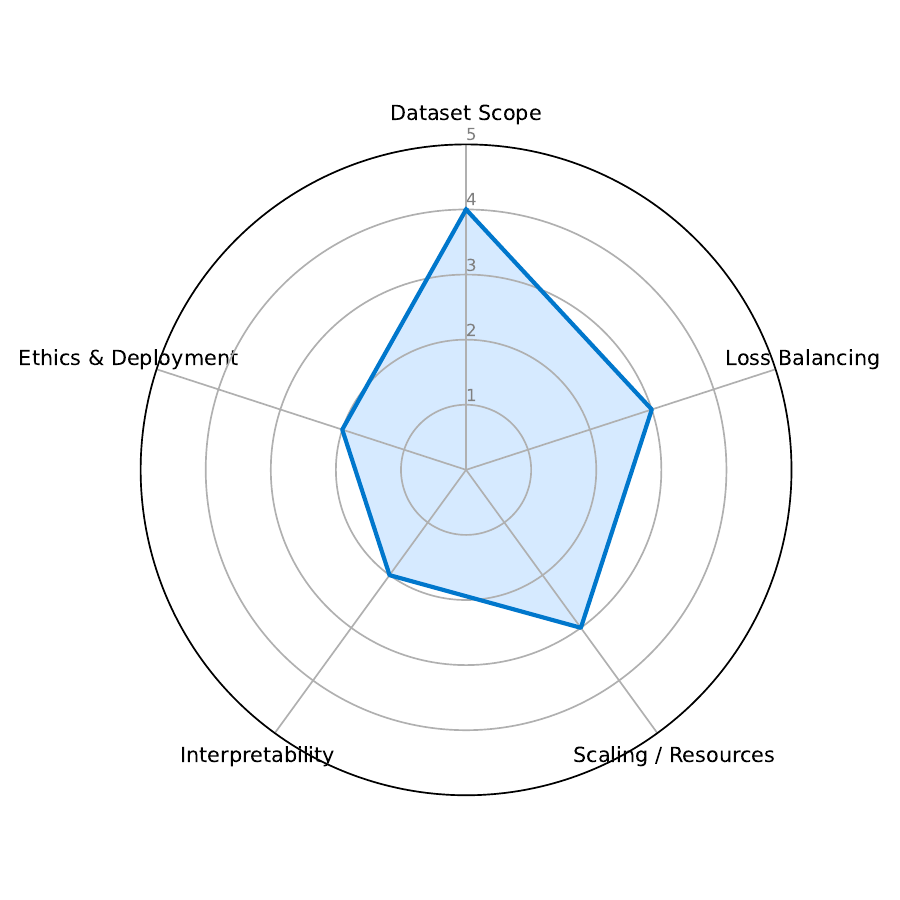}
  \caption{Key limitation dimensions of TRISKELION-1:
  dataset scope, loss balancing, scaling, interpretability, and ethics.}
  \label{fig:p5_limitations_chart}
\end{figure}
\section{Future Work}
\subsection{Efficient Scaling}
\begin{itemize}
  \item Extend to sparse Mixture-of-Experts (MoE) and adaptive routing architectures.
  \item Apply quantization and distillation to compact unified D–P–G models.
  \item Investigate empirical scaling laws governing joint-loss convergence.
\end{itemize}

\subsection{Neurosymbolic Integration}
\begin{itemize}
  \item Combine symbolic reasoning with deep generative components
        for causal interpretability.
  \item Employ knowledge graphs to constrain latent associations
        in safety-critical domains.
\end{itemize}

\subsection{Interpretability and Trust}
\begin{itemize}
  \item Define standardized metrics for descriptive interpretability
        (e.g., mutual information and cluster coherence).
  \item Conduct user studies in scientific, industrial,
        and educational settings where transparency is vital.
\end{itemize}
\subsection{Safety and Ethical Scaling}
\begin{itemize}
  \item Incorporate adversarial robustness and privacy-preserving learning.
  \item Establish audit pipelines for fairness and bias testing.
  \item Explore federated training to preserve data confidentiality.
\end{itemize}
\begin{figure}[H]
  \centering
  \includegraphics[width=\linewidth]{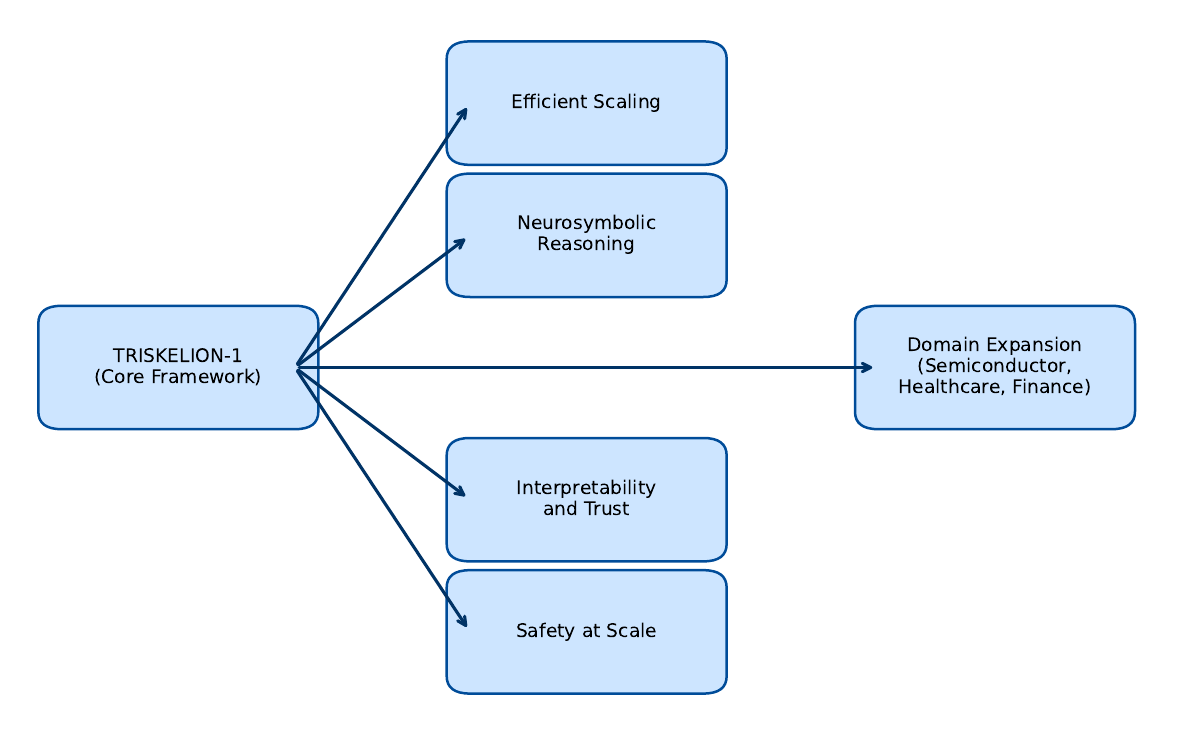}
  \caption{High-level roadmap illustrating the planned evolution of TRISKELION-1 
  from the core framework toward five research directions: efficient scaling, 
  neurosymbolic reasoning, interpretability and trust, safety at scale, 
  and domain expansion into applied sectors such as semiconductors, healthcare, 
  and finance.}
  \label{fig:p5_futurework_map}
\end{figure}
\subsection{Application Domains}
\begin{itemize}
  \item \textbf{Semiconductors:}
        Combine process, metrology, and inspection data for intelligent yield learning and digital-twin optimization.
  \item \textbf{Healthcare:}
        Apply TRISKELION-1 to multimodal medical datasets
        (ECG, MRI, genomics) for interpretable diagnostics.
  \item \textbf{Finance:}
        Employ unified predictive–generative coupling for
        anomaly detection and time-series forecasting.
\end{itemize}
\FloatBarrier
\vspace{-0.8em}
\section{Closing Remarks}
From conception to experimental validation, \textbf{TRISKELION-1} demonstrates that descriptive, predictive, and generative
intelligence can coexist harmoniously within a unified latent substrate.

\begin{figure}[H]
  \centering
  \includegraphics[width=\linewidth]{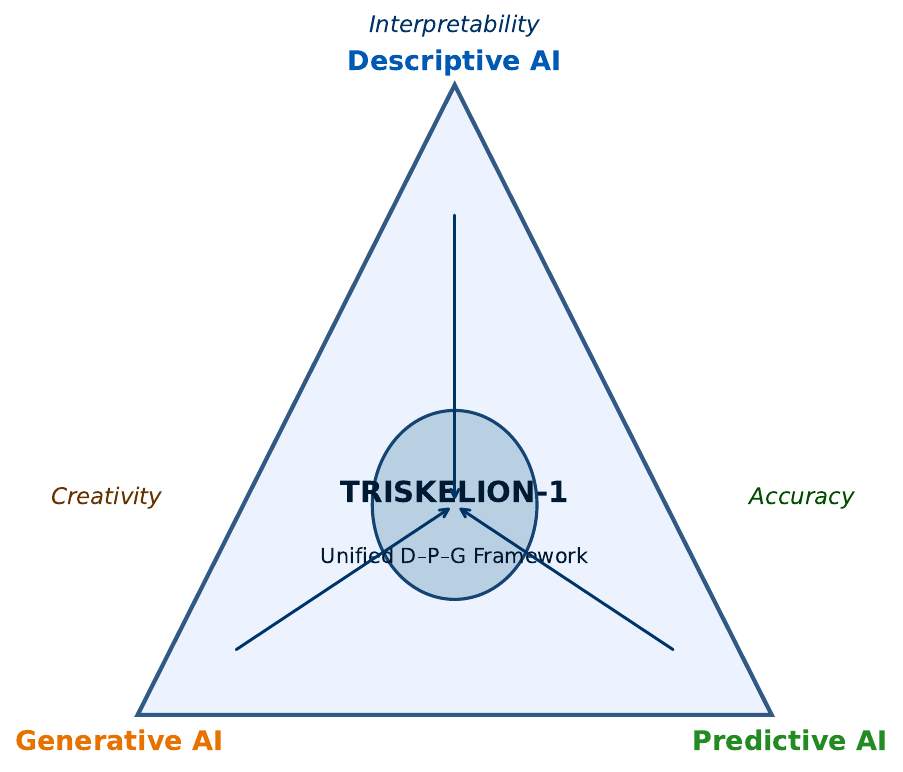}
  \caption{Holistic AI vision: TRISKELION-1 as the convergence point of descriptive, predictive, and generative paradigms—balancing interpretability, accuracy, and creativity within a unified latent framework.}
  \label{fig:p5_holistic_vision}
\end{figure}

Finally, Fig.~\ref{fig:p5_holistic_vision} summarizes the overarching vision of
\textbf{TRISKELION-1} as a unified Descriptive–Predictive–Generative framework.
The diagram conceptually situates the model at the intersection of the three
AI paradigms—descriptive, predictive, and generative—illustrating how unified
optimization enables a balance between interpretability, accuracy, and
creativity.  
This holistic representation captures the ultimate goal of the framework:
to integrate understanding, forecasting, and creation within a single latent
architecture that generalizes across modalities and domains (see
Fig.~\ref{fig:p5_holistic_vision}).
This work establishes a foundation for \textit{holistic AI}—
systems that \emph{describe}, \emph{predict}, and \emph{generate}
within the same cognitive and computational framework.
Future research (see Section~\ref{sec:part4}) can extend these findings
toward multimodal, real-world, and ethically aligned intelligent systems.
\appendices
\section{Extended Proofs and Conceptual Analysis}
\label{app:theory}
\subsection{Mutual-Information Preservation}
For completeness, we restate the manifold-alignment principle used in
the descriptive regularizer. Let
\[
L_{\text{manifold}}
  = \sum_{(i,j)}\!\left[
     w_{ij}\log q_{ij} + (1-w_{ij})\log(1-q_{ij})
    \right],
\]
where \(w_{ij}\) and \(q_{ij}\) encode neighborhood affinities in
high- and low-dimensional spaces.
Define binary variables \(U,V\) indicating neighbor membership with
\(P(U{=}1)=w_{ij}\) and \(P(V{=}1)=q_{ij}\).
The cross-entropy satisfies
\(H(U,V)=H(U)+D_{\mathrm{KL}}(U\!\|\!V)\),
so minimizing it reduces \(D_{\mathrm{KL}}\) and increases a
lower bound on mutual information,
\[
I(U;V)\ge H(U)-H(U,V).
\]
Hence manifold-preserving losses implicitly maximize local
mutual-information alignment between original and embedded manifolds.
\begin{figure}[H]
  \centering
  \includegraphics{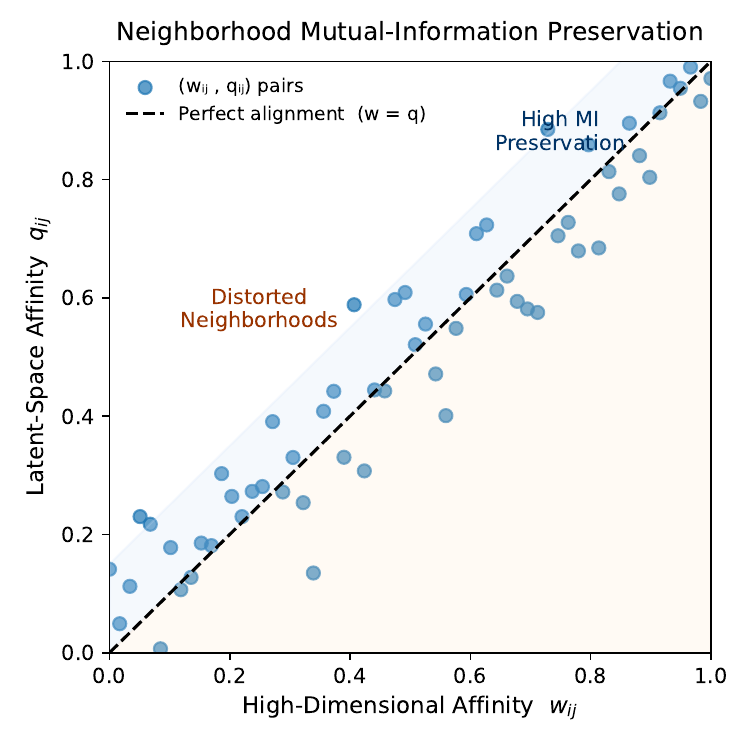}
  \caption{Neighborhood mutual-information preservation between high-dimensional affinities \(w_{ij}\) and latent-space affinities \(q_{ij}\). Points near the diagonal indicate faithful manifold alignment, while deviations reflect local distortion.}
  \label{fig:appA_mi_preservation}
\end{figure}
\subsection{Calibration under Cluster Regularization}
For predictor \(f_\phi\) with latent \(z\) and clustering
\(C(z)\), the regularized risk is
\[
R(f)=
\mathbb{E}[\ell(f(z),y)]
+\lambda D_{\mathrm{KL}}\!\big(q(z)\,\|\,C(z)\big).
\]
If \(\ell\) is classification-calibrated, the Bayes-optimal predictor
\(f^{\star}(z)
=\arg\min_f \mathbb{E}[\ell(f(z),y)\mid z]\)
remains unchanged for sufficiently small~\(\lambda\)
when clusters respect label partitions.
Thus the regularizer biases boundaries toward cluster-consistent
decisions while preserving calibration and accuracy.
\begin{figure}[H]
  \centering
  \includegraphics{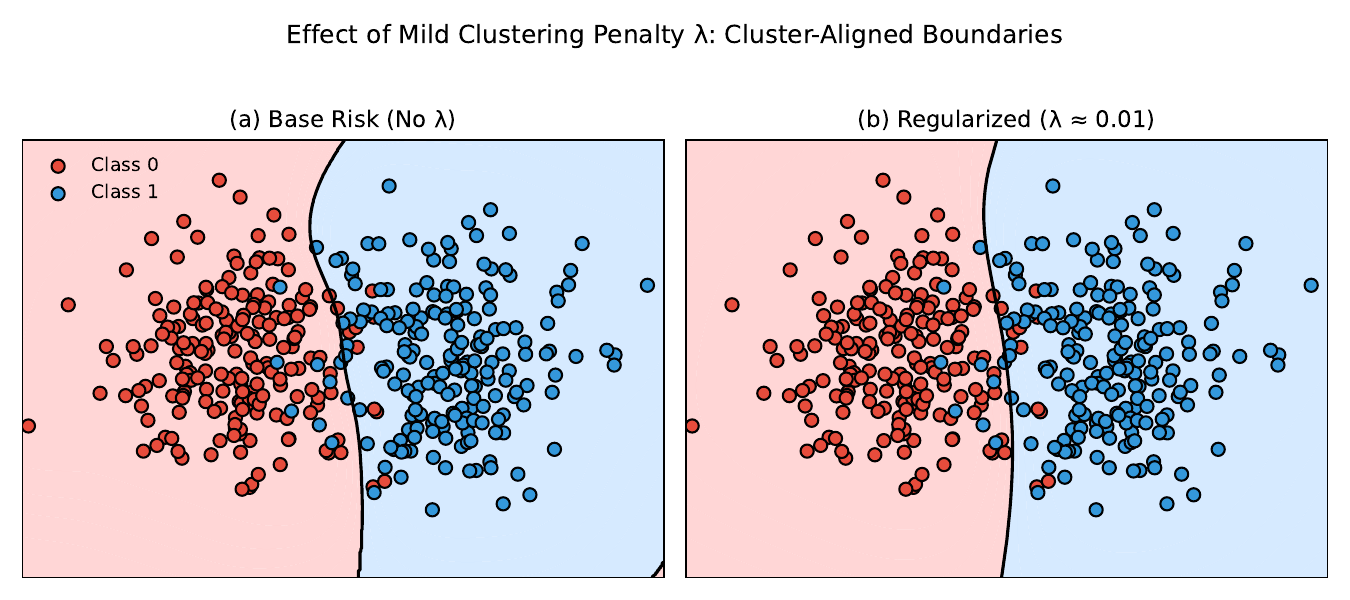}
  \caption{Effect of a mild clustering penalty \(\lambda\).
  Calibration is preserved while the decision boundary becomes smoother and aligns with cluster geometry. Left: unregularized classifier; Right: \(\lambda\)-regularized classifier respecting latent structure.}
  \label{fig:appA_calibration}
\end{figure}
\subsection{Rademacher-Complexity Reduction via Smoothness}
For hypothesis class \(\mathcal{H}\), the empirical Rademacher
complexity is
\[
\hat{\mathcal{R}}_S(\mathcal{H})
=\frac1N\mathbb{E}_\sigma\!\left[
 \sup_{h\in\mathcal{H}}\sum_i\sigma_i h(x_i)
 \right].
\]
A Laplacian penalty
\(\Omega(z)=\sum_{(i,j)} w_{ij}\|z_i-z_j\|^2\)
enforces smoothness across neighboring samples.
With Laplacian \(L_w=U\Lambda U^\top\),
the effective capacity scales as
\[
\hat{\mathcal{R}}_S(\mathcal{H}\!\circ\!E_\theta)
 \le
 \tilde{\mathcal{O}}\!\left(
 \frac{\mathrm{Lip}(f)\,\mathrm{Lip}(E_\theta)}
      {\sqrt{N}}
 \sqrt{\lambda_{\min}^{-1}(L_w)}
 \right).
\]
Hence manifold regularization reduces hypothesis complexity and
improves generalization in the TRISKELION-1 encoder.
\begin{figure}[H]
  \centering
  \includegraphics{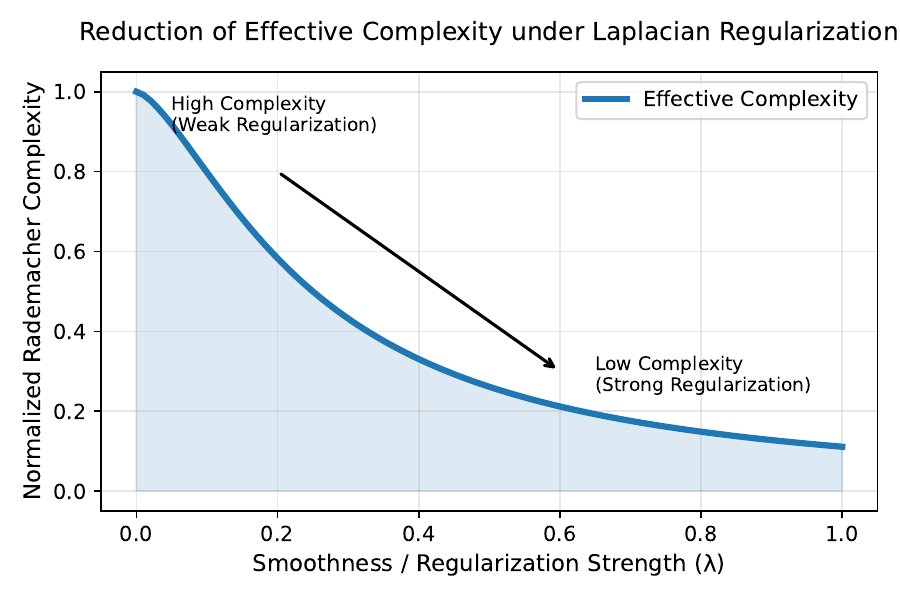}
  \caption{Conceptual reduction of effective Rademacher complexity under Laplacian-smoothness regularization.
Increasing the smoothness coefficient \(\lambda\) reduces hypothesis capacity, yielding smoother decision boundaries and improved generalization.}
  \label{fig:appA_rademacher}
\end{figure}
\subsection{Cross-Gradient Coupling and Stability}
\begin{figure}[H]
  \centering
  \includegraphics{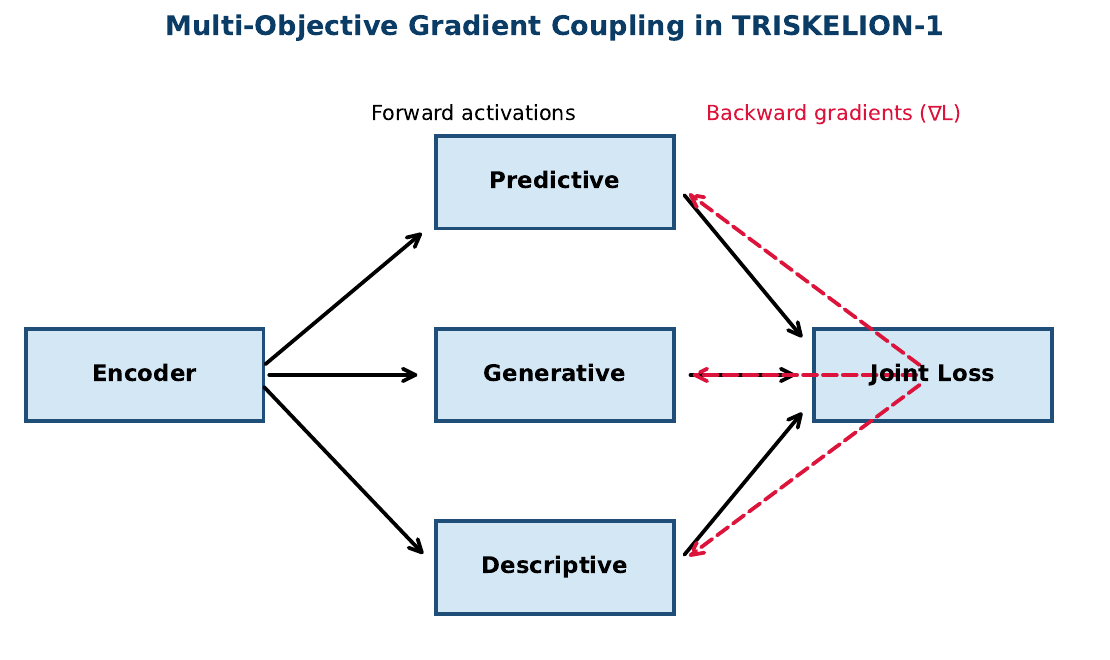}
  \caption{Schematic of multi-objective gradient coupling in TRISKELION-1.
Solid black arrows indicate forward activations through predictive, generative, and descriptive branches.
Red dashed arrows denote back-propagated gradients (\(\nabla L\)) from the joint loss, combining weighted components (\(\alpha,\beta,\gamma\)) to update the shared encoder.}
  \label{fig:appA_gradients}
\end{figure}

Let the unified loss
\(\mathcal{L}_{\text{TRI}}
  = \alpha\mathcal{L}_{\text{pred}}
  + \beta\mathcal{L}_{\text{gen}}
  + \gamma\mathcal{L}_{\text{desc}}\).
During training, the encoder update follows
\[
\nabla_\theta \mathcal{L}_{\text{TRI}}
= \alpha\nabla_\theta \mathcal{L}_{\text{pred}}
 + \beta\nabla_\theta \mathcal{L}_{\text{gen}}
 + \gamma\nabla_\theta \mathcal{L}_{\text{desc}}.
\]
Empirically, these gradient directions remain partially aligned;
hence optimization behaves as a
Pareto-stable multi-objective descent.
Figure~\ref{fig:appA_gradients} visualizes this coupling.
\section{Complexity Analysis}
\label{app:complexity}
\noindent
Table~\ref{tab:appB_complexity} summarizes the approximate analytical
scaling behavior of the main TRISKELION-1 components.
The goal is not to report measured runtimes but to provide
a theoretical sense of how computation and memory grow with
latent dimensionality \(m\), batch size \(N\), and class count \(C\).
\begin{table}[H]
\centering
\caption{Approximate computational characteristics of principal TRISKELION-1 modules (per sample).}
\resizebox{\columnwidth}{!}{%
\begin{tabular}{lccc}
\toprule
\textbf{Module} & \textbf{FLOPs} & \textbf{Memory} & \textbf{Scaling Law} \\
\midrule
Encoder (CNN)            & $\mathcal{O}(m^2)$ & $\mathcal{O}(m)$ & Quadratic in feature maps \\
Predictive head          & $\mathcal{O}(mC)$  & $\mathcal{O}(C)$ & Linear in classes $C$ \\
Generative decoder (VAE) & $\mathcal{O}(m^2)$ & $\mathcal{O}(m)$ & Quadratic in latent size \\
Descriptive regularizer  & $\mathcal{O}(Nm)$  & $\mathcal{O}(m)$ & Linear in batch size $N$ \\
Joint loss aggregation   & $\mathcal{O}(m)$   & negligible       & Linear combination \\
\bottomrule
\end{tabular}%
}
\label{tab:appB_complexity}
\end{table}
\noindent
These estimates highlight that encoder and generative operations
grow quadratically with latent dimensionality,
whereas predictive and descriptive terms scale linearly with
either the number of classes or batch size.
They serve as analytical guidelines illustrating relative
scaling among modules—no empirical profiling was performed.
\section{Notation and Assumptions}
\label{app:notation}
\paragraph{Core Symbols.}
\begin{itemize}
  \item $\mathbf{x}$ — input sample
  \item $\mathbf{z}$ — shared latent embedding
  \item $E_\theta$ — encoder network
  \item $f_\phi$ — predictive head
  \item $p_\psi$ — generative decoder
  \item $C,\,R,\,T$ — descriptive modules (clustering, reduction, topic)
  \item $\mathcal{L}_{\text{pred}},\,
         \mathcal{L}_{\text{gen}},\,
         \mathcal{L}_{\text{desc}}$ — loss components
  \item $(\alpha,\beta,\gamma)$ — weighting coefficients
  \item $L_w$ — graph Laplacian for smoothness regularization
\end{itemize}
\paragraph{Assumptions.}
\begin{itemize}
  \item All activations are Lipschitz-continuous (ReLU or GELU).  
  \item Noise processes are Gaussian unless otherwise noted.  
  \item Training data are finite and i.i.d. samples from
        \(p_{\text{data}}\).  
  \item Loss weights \((\alpha,\beta,\gamma)\) are tuned empirically
        but can be optimized via hyper-search.  
\end{itemize}


\balance
\bibliographystyle{IEEEtran}
\bibliography{references}

\end{document}